\documentclass[preprint,12pt]{elsarticle}
\usepackage{amssymb}
\usepackage{url}
\usepackage{hyperref}
\journal{IEEE Access and accepted in this journal}
\usepackage{graphicx}
\usepackage{svg}
\usepackage{amsmath}
\usepackage{slashbox}
\usepackage{algorithm}
\usepackage{algpseudocode}

\DeclareMathAlphabet      {\mathbfit}{OML}{cmm}{b}{it}
\begin{document}

\begin{frontmatter}



\title{Npix2Cpix: A GAN-based Image-to-Image
Translation Network with Retrieval-Classification Integration for Watermark Retrieval from Historical Document Images}


\author[inst1,inst2]{Utsab Saha}

\affiliation[inst1]{organization={Department of Electrical and Electronic Engineering},
            addressline={BUET}, 
            city={Dhaka},
            postcode={1000}, 
            country={Bangladesh}}

\author[inst3]{Sawradip Saha}
\author[inst1]{Shaikh Anowarul Fattah}
\author[inst4]{Mohammad Saquib}

\affiliation[inst2]{organization={Department of Computer Science and Engineering},
            addressline={BRAC University}, 
            city={Dhaka},
            postcode={1212}, 
            country={Bangladesh}}
\affiliation[inst3]{organization={Department of Mechanical Engineering},
            addressline={BUET}, 
            city={Dhaka},
            postcode={1000}, 
            country={Bangladesh}}
\affiliation[inst4]{organization={Department of Electrical Engineering},
            addressline={The University of Texas at Dallas}, 
            city={Richardson},
            postcode={TX 75080}, 
            country={United States}}

\begin{abstract}
The identification and restoration of ancient watermarks have long been a major topic in codicology and history. Classifying historical documents based on watermarks is challenging due to their diversity, noisy samples, multiple representation modes, and minor distinctions between classes and intra-class variations. This paper proposes a modified U-net-based conditional generative adversarial network (GAN) named Npix2Cpix to translate noisy raw historical watermarked images into clean, handwriting-free watermarked images by performing image translation from degraded (noisy) pixels to clean pixels. Using image-to-image translation and adversarial learning, the network creates clutter-free images for watermark restoration and categorization. The generator and discriminator of the proposed GAN are trained using two separate loss functions, each based on the distance between images, to learn the mapping from the input noisy image to the output clean image. After using the proposed GAN to pre-process noisy watermarked images, Siamese-based one-shot learning is employed for watermark classification. Experimental results on a large-scale historical watermark dataset demonstrate that cleaning the noisy watermarked images can help to achieve high one-shot classification accuracy. The qualitative and quantitative evaluation of the retrieved watermarked image highlights the effectiveness of the proposed approach.
\end{abstract}




\end{frontmatter}


\section{Introduction}
\label{sec:introduction}
Watermark recognition from historical images and retrieval is a crucial task for historians and archivists in order to classify and explore the origin of a historical document. Ancient watermarks help the users to estimate the dates of creation of the watermarked documents (texts or drawings) without any explicit chronological indication \cite{sasaki1992dating} \cite{teodonio2022late}, as well as to determine the probable location of the document's origin \cite{heawood1924use}. For archivists, historians, curators, and anyone else interested in the trade of written documents, being able to recognize the watermark on historical papers and obtain the clean version of that watermark, makes it possible to precisely date the paper and determine the paper manufacturer by looking through existing watermark design catalogs (booksellers, antiquarians, etc.). Due to natural deterioration as well as man-made catastrophes, there is a danger that the information contained in the old documents would be lost after many years. Even if the documents are preserved carefully, there is a great possibility that they will continue to deteriorate over time. Ancient watermarks are essential for gathering crucial data, and categorizing historical documents. That is why, in recent years, it has become more essential than ever to be able to accurately identify historical watermarks in documents. Despite its great usefulness, automatic watermark detection remains highly challenging and remains unresolved. In addition to being a useful resource in the preservation of cultural heritage collections, historical watermarks serve as a source of information for archivists and historians. Due to the variety of patterns used in these watermarks, they are very difficult to identify and recognize. The identification issue is further complicated by the presence of handwriting on the watermarks and various types of noises \cite{habibunnisha2019reduction}.

Hand tracing and backlit photography are two traditional methods of watermark retrieval \cite{boyle2009watermark}. A number of research works have emphasized localizing and extracting a watermark pattern from a back-lit image, rarely incorporating  aligned reflected light images \cite{hiary2007system}, 
\cite{hiary2008paper}, \cite{said2016watermark}. 
Aside from analog watermarks on historical documents, digital watermarking became popular in the late 1990s in order to embed robust digital marks into diverse materials and images to identify the origin, owner, rights, and integrity, as well as to ensure security. In addition to steganography, spread spectrum communications technology, and noise theory-based generic watermarking systems \cite{JIANZHAO1998397}, there has been a lot of progress in computer graphics-based robust watermarking tools, such as block-based \cite{DARMSTAEDTER1998417} and triangle-based ones \cite{YIN2001409}. Automatic image retrieval has long been recognized as an intriguing method for historical document study. In the late 1990s, Rauber et al. \cite{Rauber1996ArchivalAR} and Belov et al. \cite{belov1999physical} discussed watermark retrieval as a use case for content-based image retrieval (CBIR) approaches. A CBIR system originally developed for trademark retrieval has been applied to tracings and radiographies of historical watermarks in the Northumbria watermarks archive project \cite{brown2002images}. Compared to radiographs, the system performed significantly better on tracings \cite{riley2003content}. Two aspects associated with document enhancement are highlighted in \cite{DE_GAN}, namely the recovery of damaged documents and watermark removal. In \cite{souibguiconditional}, many of the existing documents are digitized using smartphone cameras. These are highly vulnerable to capturing distortions (perspective angle, shadow, blur, warping, etc.), making them hard to read by a human or by an OCR engine.

Lucene Image REtrieval (LIRE) is a Java open-source program that was first made available in 2006 \cite{marques2012visual}. Based on the Lucene search engine, in the LIRE program, both global and local features are used. Low-level color, texture, and certain mixed characteristics make up the global features. Image retrieval performance is demonstrated by using the global features of LIRE's free demo edition (Fuzzy color, texture histogram, and MPEG7 color layout). Recently in \cite{wang2022interactive}, considering as an application of the proposed method,  digital watermarks from images have been eliminated. Initially, watermark-related nodes are chosen using the various schemes proposed here. Finally, local manipulation is employed to delete all associated control points with the watermark. In \cite{yuan2021blind}, a blind color picture watermarking system with good spatial performance is proposed to achieve effective copyright protection of color images by combining the advantages of spatial-domain and frequency-domain watermarking schemes. Image watermarking algorithms such as 1D empirical mode decomposition (EMD) and dimensional reduction via Hilbert curve \cite{hu2022robust}, watermarking system with a memristor-based hyperchaotic oscillator \cite{sehra2022secure}, two-stage fuzzy inference system \cite{gong2022visible}, and others have recently gained popularity for the purpose of protecting copyright and preventing information from being used without authorization.

Deep learning algorithms are getting more attention for historical document analysis. Identifying an abstract representation—such as drawings, paintings, plush toys, sculptures, signs, or silhouettes—is an easy and frequent task for humans, but computers continue to struggle with it. Histogram-based descriptors \cite{belongie2000shape} have been used in older studies like \cite{rauber1997retrieval}, whereas the use of machine learning techniques like dictionary learning \cite{picard2016non} and neural networks \cite{pondenkandath2018identifying} are noticeable in more recent works.
In \cite{picard2016non}, a retrieval strategy for categorizing watermarks is proposed based on local features and a dictionary that is trained on a large amount of data. The quest for comparable deep learning-based historical watermark identification and recognition systems has been made by several researchers.  The technique provided in \cite{pondenkandath2018identifying}, solves the problem of identifying the same object even when it is depicted in a variety of manners also known as the cross-depiction issue, on a historical watermark dataset \cite{frauenknecht2015wzis}.  This dataset is not publicly available and contains 106,000 watermark pictures. 

The research landscape in historical document analysis has seen significant advancements, particularly in the domains of segmentation and recognition \cite{ma2020segmentation}. The method for text baseline detection from historical documents has been established \cite{jia2021detecting}, while handwritten text recognition (HTR) in historical documents has witnessed remarkable progress through the application of deep learning techniques \cite{aradillas2021boosting}. A noteworthy contribution is the introduction of a Human-Inspired Recognition system for pre-modern Japanese historical documents \cite{le2019human}, along with the proposal of a character detection and segmentation method for historical Uchen Tibetan documents \cite{zhang2022character}. A newer approach is demonstrated in \cite{festa2023ancient}, which combines X-ray fluorescence and Fourier transform infrared spectroscopies along with machine learning to gain insights into the genesis of an ancient manuscript. These recent works underscore the crucial role of machine learning and deep learning in extracting precise information about specific cultures from historical data. The incorporation of historical watermarks, frequently encountered in ancient official documents, emerges as a significant asset in this pursuit.
 
Recently, research on image retrieval for watermarks has been jointly conducted by the Institut de Recherche et d'Histoire des Textes (IRHT), the Institut National de Recherche en Informatique et en Automatique (INRIA), and the Ecole Nationale des Chartes and results are reported in \cite{shen2021large}. A large public dataset with over 6000 original images is published here, allowing researchers to tackle scenarios of practical importance at scale. A matching score and feature fine-tuning technique based on filtering local matches is also introduced here using spatial consistency. In \cite{bounou2020web}, a new publicly accessible online tool for automated watermark identification is proposed based on \cite{shen2021large}. In \cite{shen2022spatially}, a spatially consistent feature matching and learning algorithm is proposed for historical image analysis. Here, an image similarity score based on geometric validation of mid-level features is defined. In this work, it is also demonstrated how weak or no supervision may be used to fine-tune out-of-the-box features for the target dataset using spatial consistency. Italian papers of the late Middle Ages with watermarks are classified using machine learning in \cite{teodonio2022late}. Here, a new spatial-temporal methodology for tracing regional paper mill production and regional recipes is also introduced. Another issue with watermark detection or classification is the multi-domain representations of data, that can be solved using recent studies like \cite{gao2022multi}. Recently, in \cite{mypaper}, a Siamese-based one-shot network for automatic watermark recognition from the historical image dataset has been proposed. Here, a thresholding-based image enhancement technique is employed, that strongly depends on the chosen threshold value. The introduction of the one-shot classification in historical watermark classification, proposed in this paper, is very useful for the low data regime, which has great demand. 

Reviewing the existing studies it can be agreed that the common problem in watermark detection is the scarcity of clean examples to test newer watermarks. There are also a lot of messy and distorted samples. These watermarks can be challenging to detect at times because of the multiple modes of representation, small discrepancies between classes, and substantial intra-class variations. As a solution to such problems, in this paper, a conditional generative adversarial network (GAN), named Npix2Cpix, is proposed to translate raw, noisy images to their cleaned counterparts. Although deep learning-based models have been used previously to remove entities from images \cite{sharma2021deep}, \cite{matsui2020gan} image-to-image translation applications to extract a specific piece of information or pattern from historical document images have not yet been explored. A recent approach improves multimodal image-to-image translation by using a shared encoder and generator for all domains, enhancing feature extraction and performance in complex tasks \cite{9811405}. Similarly, another method enhances change detection in optical and SAR images by aligning them into a common feature space, achieving higher accuracy and robustness \cite{li2021deep}. 

The phrase "translating" refers to converting an input noisy image into a corresponding clean output image, that can be used to describe a variety of issues in image processing, computer graphics, and computer vision. A U-net-based architecture serves as the generator of the GAN, that has been proposed here, reconstructing the clear image from the noisy one. Both the generated image and the noisy watermarked image are compared with the instance's ground truth. The generator is trained to produce images that are somewhat similar to the ground truth, while the discriminator is taught to discern between ground truth and generated images. To update the parameters of the network, a combined loss function is employed. Subsequently, for the categorization of historical watermarks, a Siamese-based one-shot classification pipeline is proposed. Here, two identical neural networks are used to process both support and query images, using the same structure, weights, and parameters. Cosine similarity is used in this case to calculate similarity.
The main contributions of this paper are:
\begin{itemize}
  \item A modified U-net-based generative adversarial image-to-image translation network for watermark retrieval from historical documents.
  \item A Siamese-based one-shot approach for classifying the retrieved watermarks. 
  
\end{itemize}
The proposed U-Net-based conditional GAN is not simply used to generate new synthetic instances of images, rather it focuses on enhancing the clarity and quality of watermarked historical document images, facilitating subsequent watermark classification and analysis. Hence the generator network of the proposed U-Net-based GAN is specially designed to map a noisy watermarked image to a generated noise-reduced watermarked image. In particular, different modifications are incorporated in the U-Net structure used in the generator network, such as an increased number of up-sampling and down-sampling blocks, a dropout layer in the up-sampling path, and the Tanh activation function in the final layer. In the proposed Siamese-based one-shot classification network, various pretrained models, including ResNet, MobileNet, and EfficientNet, are integrated as feature extractors, along with cosine distance-based similarity measurements. This approach is designed to handle extremely degraded historical documents that may not be fully denoised, distinguishing it from traditional Siamese networks.
By combining the modified U-Net-based conditional GAN for noisy image to noise-reduced image translation and a subsequent one-shot classification scheme using the proposed Siamese network, the proposed method effectively addresses the complex challenges of classifying ancient watermarked document images.

The next sections will assist in developing an understanding of the dataset used in this work, our proposed approach, results, and analysis on the comparison of results with baselines. 
\begin{figure*}[h!]
    \centering
    \includegraphics[width=.9\linewidth]{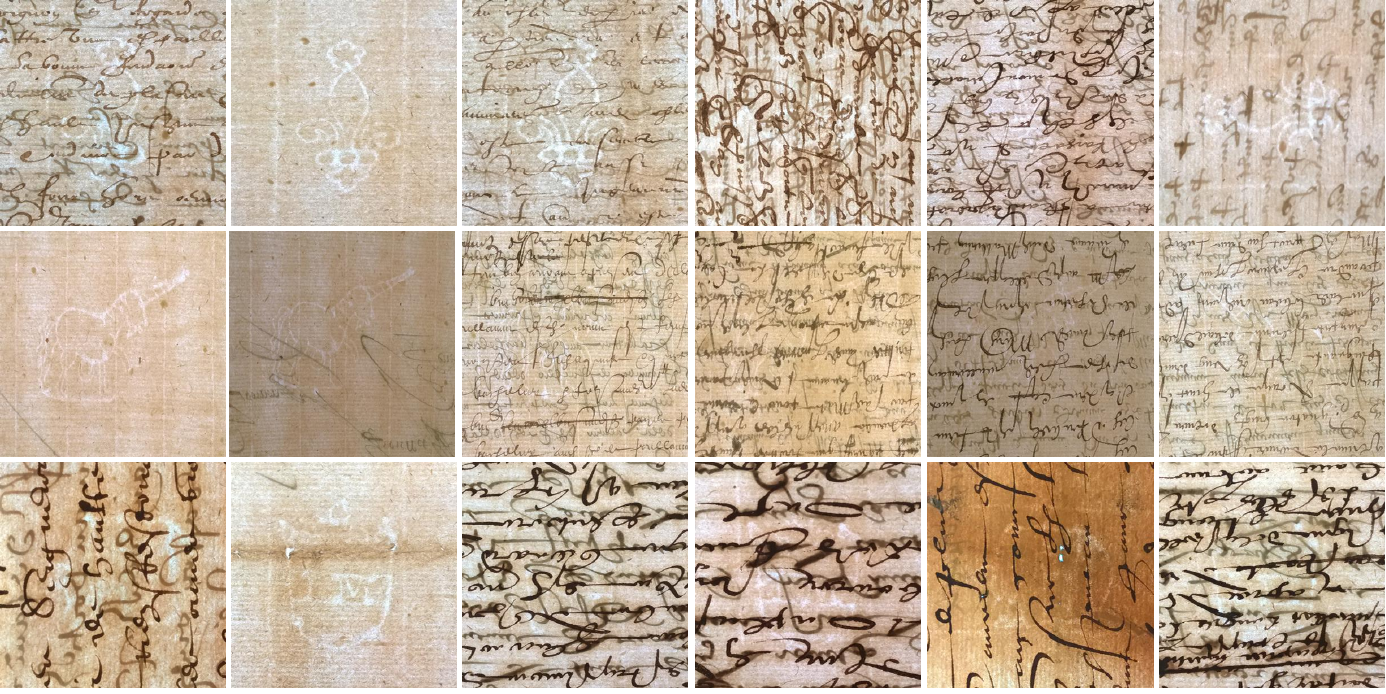}
    \caption{Noisy watermarked images containing watermarks in different backgrounds with hazy hand writings}
    \label{fig: raw_watermarks}
\end{figure*} 
\begin{figure*}[h!]
    \centering
    \includegraphics[width=0.9\linewidth]{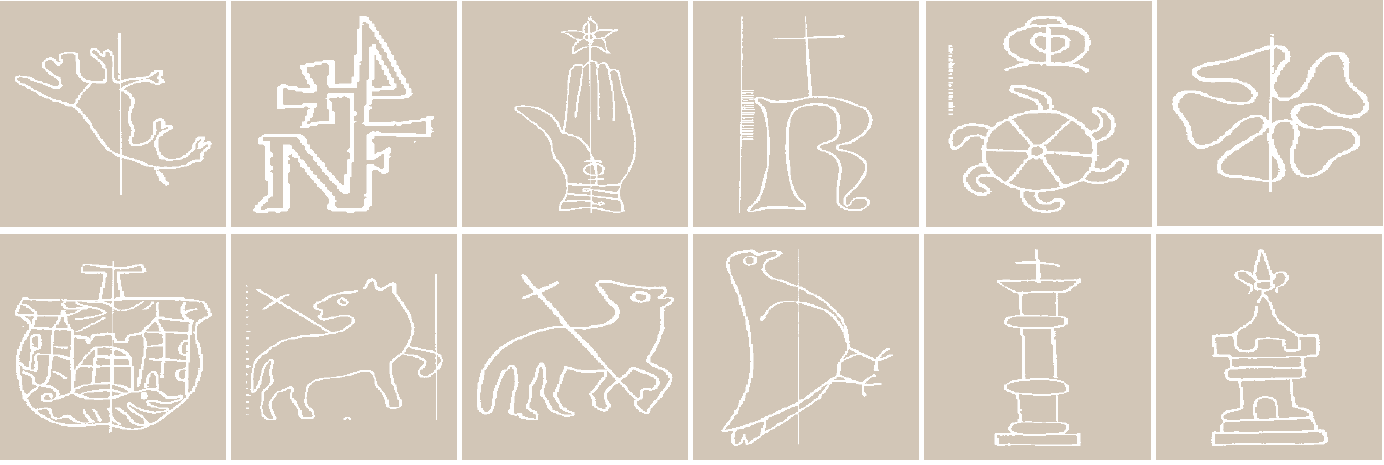}
    \caption{Synthetic watermarks from different classes}
    \label{fig: Syn_watermarks}
\end{figure*}
\section{Dataset Description}\label{sec2}

Real-world data collecting or the accessibility of public datasets is one of the distinctive features of research. A large-scale, diverse watermark dataset with a significant number of samples for each fine-grained class is extremely difficult to come across. The dataset used in this work  \cite{shen2021large}, which has a large arrangement and variety compared to other historical watermark recognition research studies that use comparably smaller datasets, gives it an advantage when it comes to training our suggested deep learning model. This dataset includes both crowded and noisy pictures and has a large number of well-defined categories. The historical watermarks on the photographs additionally make it challenging to identify patterns because of the images' significant intra-class diversity and minimal between-class differentiation. There are more than 16,000 400x400-pixel images. Each of the dataset's several parts focuses on a different objective. The "Classification" dataset and the "Briquet synthetic" dataset are segments taken into account in this work. The objective of the "Classification" dataset, as shown in Fig.\ref{fig: raw_watermarks},  is the development and evaluation of algorithms for watermark retrieval and one-shot, fine-grained watermark classification from images. We divide this dataset into two parts for ease of use: the first part (60\%) is for feature/meta-training of the Siamese network and includes several instances of each watermark class, while the second part (40\%) is for testing one-shot identification. To begin, all the images in the classification dataset (50 images per class; a total of 100 classes) are pre-processed with a GAN-based watermark retrieval network to extract the precise watermarks. A Siamese-based one-shot classification is carried out to evaluate if the watermarks are correctly retrieved. 

Furthermore, in "Briquet synthetic" dataset, synthetic images are created from the artworks by highlighting the drawing style and using the background color of a typical watermark. In this dataset, 16,753 synthetic drawings are present as shown in Fig. \ref{fig: Syn_watermarks}.
\section{Methodology}\label{sec3}

\begin{figure*}[h!]
    \centering
    \includegraphics[width=1\textwidth]{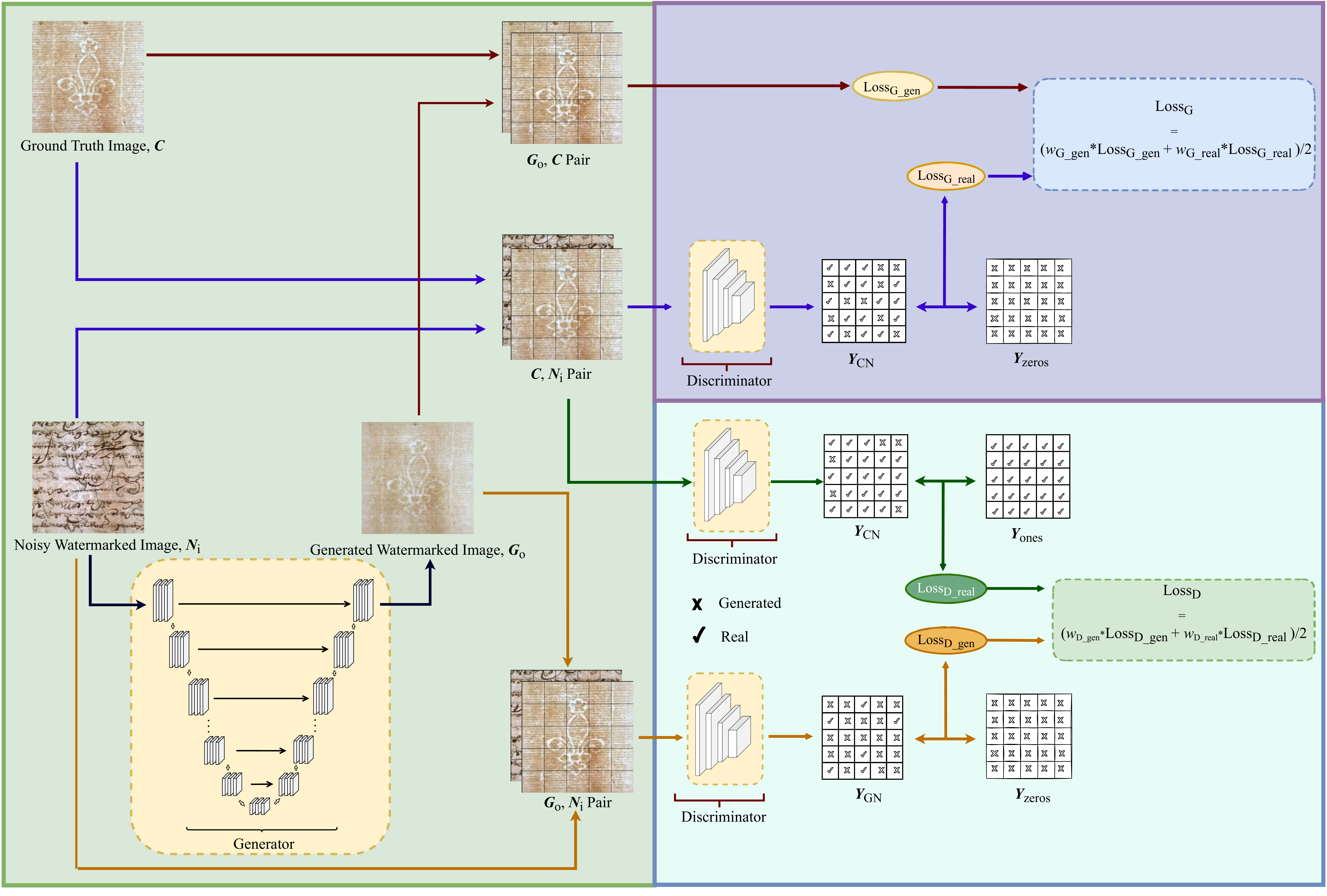}
    \caption{Flowchart of the proposed U-net-based GAN}
    \label{main_proposed_flowchart}
\end{figure*}
Traditional methods for watermark retrieval and classification usually input the noisy image directly into the classifier network, which then does classification, and recent works for denoising documents and classification are based on deep residual networks and auto-encoders \cite{chen2017study}, \cite{neji2019adversarial}. As a result, additional effort is necessary to expend into developing a robust classification network. This is because, in this case, the classification network will have to deal with both the classification and noise reduction problems. Thus, the model could not provide sufficient performance, especially if the input image is severely degraded. Other techniques focus specifically on the watermarked regions for retrieval. As ancient documents with watermarks contain handwriting and additional noise, there is a possibility that it ends up enhancing the noise and writing of that region as well. On the other hand, a crucial part of this research is to handle one-shot classification where there is less scope for getting a huge amount of data for traditional classification. As a solution to the problems associated with traditional methods, this research introduces two major schemes. Initially, in an effort to mitigate the impact of noise and handwritten artifacts, a U-net-based conditional GAN is proposed in this paper. This network facilitates the transformation of noisy historical images into their clean counterparts by effectively eliminating handwriting and extraneous noise, thereby constituting an image-to-image translation scheme. In order to accomplish noise removal and watermark extraction, the suggested image-to-image translation approach with U-net-based GAN has been named Npix2Cpix. Furthermore, to address the limitation imposed by the scarcity of training data, a modified Siamese-based one-shot classification network is used. A GAN-based network employed prior to the one-shot classification stage is very efficient, as instead of a noisy image, a noise-reduced as well as writing removed image needs to be handled in the classifier. The overall idea proposed in this paper can be summarized into two phases, they are - 
noise and writing removal using U-net-based GAN and one-shot classification. The problem statement along with suggested noise and handwriting removal with classification pipelines are described in the following subsections.

\subsection{Problem Statement} 
\label{ssec:ps}
Historical images, to be more specific, historical watermarked document images usually have some additional challenges compared to traditional computer vision tasks. Specifically, there are three major technical challenges in the watermark retrieval from 
historical document images, namely- 
\begin{itemize}
    \item background of historical images
    \item the variation in watermarks
    \item the presence of handwriting 
\end{itemize}
The estimation of background within historical document images holds significant importance due to the presence of watermarks typically situated in the background section of the image. However, detecting the background in such instances represents a notably challenging task.
The background of historical images is often noisy, blurry, of different brightness levels, spotted, and non-homogeneous, as can be seen in Fig. \ref{fig: raw_watermarks}. Here, noise refers to the discontinuity and non-uniformity of the background paper. The degradation of watermarked documents with time, often variations in different locations of the same image as well as images from different sources, affects the brightness, clarity, sharpness, and overall colors of the watermark background. These characteristics make it difficult for the model to estimate the historical image's background. The watermarks to be retrieved from ancient document images vary significantly in terms of their types, patterns, and level of noise corruption. These watermarks can occasionally be extremely faded, distorted, and discontinuous due to the aging or data acquisition process. Besides, the patterns of the watermarks have a huge variation in their size, complexity, location, design, thickness, and clarity. However, observing images in Fig.\ref{fig: raw_watermarks}, it can be agreed that the biggest challenge to extracting significant features from these images is the blackish-brown handwriting, which is randomly spread all over most of the images. To be more specific, the handwriting on the top of the watermarks creates additional complexity in retrieving watermarks. Besides, there is a huge variation in complexity, thickness, color, depth (Dark black, Light brown, etc.), size, and style of the letters. And the biggest challenge for the model will be to generalize this variation. Besides, as the images are obtained from different sources, the handwriting not only varies in writing style and typographical characteristics, but it might also be of fonts that do not exist at present and are in no way related from image to image. These three major issues have to be handled in this research work.

\subsection{Image to Image Translation: Npix2Cpix} \label{npix2cpix}
To eliminate the handwriting from the input images as well as to remove the watermark background's discontinuity and clutter, we propose a conditional GAN-based technique for the first step of our pipeline. To put it in another way, this proposed network will receive a raw noisy watermarked image, $\mathbfit{N}_\text{i}$ as input and will generate a noise-free image, which is the generated watermarked image, $\mathbfit{G}_\text{o}$ with only the watermark. From the literature study, it is found that GANs are one of the best ways to generate a clean image from a raw noisy watermarked image. Still, it surprisingly has not yet been explored enough for the historical watermark retrieval domain. This application of GANs is vastly different from the traditional uses of GANs as well as previous watermark retrieval techniques employed in this domain. Usually, GANs are used to generate new, synthetic instances of images that can pass for real data, whereas the scope of GAN usage in this domain is limited to deblurring or removing compression artifacts. Using a conditional adversarial network, \cite{pix2pix} shows that it is efficient at a number of tasks, including colorizing pictures, reconstructing objects from edge maps, and synthesizing images from label maps. 

\begin{figure*}[h!]
    \centering
    \includegraphics[width=\textwidth]{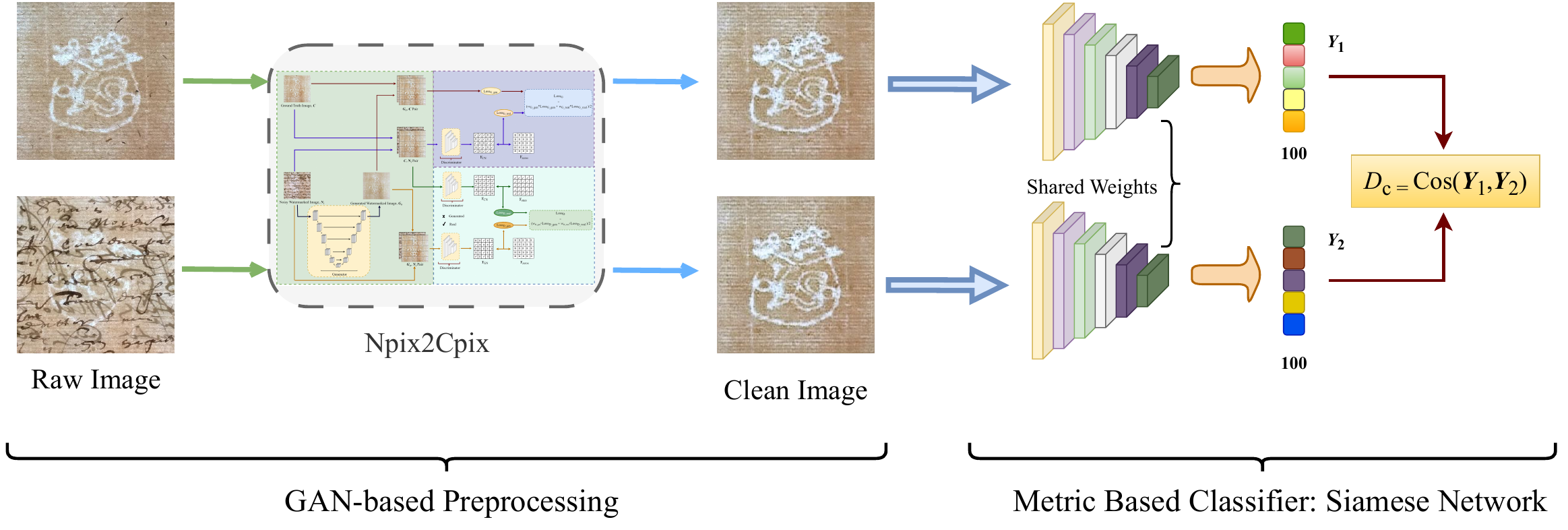}
    \caption{Flowchart of the proposed method consisting of Npix2Cpix: A U-net-based GAN with a Siamese based one-shot classification network}
    \label{fig:Siamese}
\end{figure*}

In this study, a conditional generative adversarial network (GAN) for image-to-image translation, that can produce clear watermarked images from the original noisy watermarked images, is presented. To ensure precise articulation of our ideas, we present our proposed methodology for denoising the documents as Algorithm \ref{alg:refinement_keypts}, which is depicted in Fig. \ref{main_proposed_flowchart}. Additionally, we outline the gist of the proposed one-shot learning-based training pipeline in Algorithm \ref{alg: Siamese}, visually represented in Fig. \ref{fig:Siamese}.

\begin{figure}[h]
    \centering
    \includegraphics[width=1\linewidth]{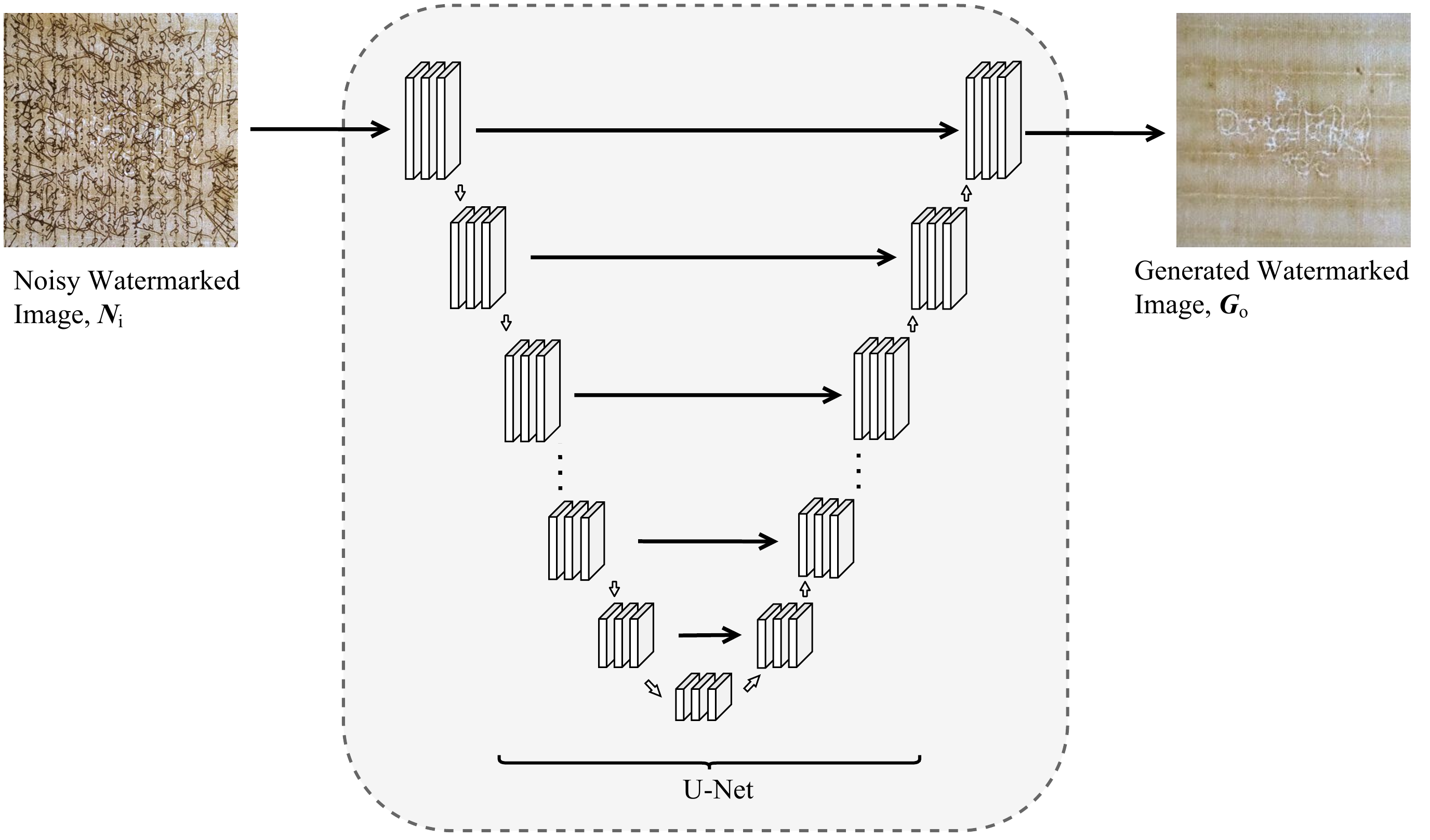}
    \caption{The generator of the proposed U-net-based GAN used for extracting clean watermarked image}
    \label{generator}
\end{figure}

Usually, a GAN consists of a generator network and a discriminator network. The generator network takes a random noise input, in order to generate target images, which is unlike our task of generating clear images from noisy images. In our case, the generator maps a high-resolution noisy watermarked image, $\mathbfit{N}_\text{i}$ to a high-resolution (clean) generated watermarked image, $\mathbfit{G}_\text{o}$ as shown in Fig. \ref{main_proposed_flowchart}. As formulated in Algorithm \ref{alg:refinement_keypts}, in order to accomplish this mapping, a noisy watermarked image, $\mathbfit{N}_\text{i}$ is passed through a generator network, containing a series of convolution blocks, along with necessary batch-normalization and pooling layers, gradually down-sampling the extracted features, finally reaching a bottleneck layer from which up-scaling of the features begins by another series of transpose convolution layers, finally obtaining the generated watermarked image, $\mathbfit{G}_\text{o}$. To give the generator a means to circumvent the bottleneck for information like this, skip connections are added, following the general shape of a U-Net \cite{unet}, which is visualized in Fig. \ref{generator}. The U-Net used for the generator offers several modifications compared to the traditional version, including an increasing number of up-sampling and down-sampling blocks, dropout in the up-sampling path to prevent overfitting, and the utilization of a Tanh activation function in the final layer. These adaptations aim to improve performance and adaptability for generating clean images from noisy watermarked document images, contrasting with conventional U-Net architectures using Softmax or Sigmoid activations primarily for segmentation tasks.

The discriminator network in a generative adversarial network has always been used as a classifier to classify the generated images as real or generated. As the case presented in this paper is more closely related to image-to-image translation, generated images should have spatial consistency with the noisy input image. Here, spacial consistency refers to the pieces of the reconstructed watermark in the generated watermarked image, $\mathbfit{G_\text{o}}$ which should be at the same locations where the corresponding parts of the original watermark in the noisy watermarked image,  $\mathbfit{N}_\text{i}$ are located. That is why this task is handled in a little different manner. The specialty of the discriminator is, unlike normal GANs, each generated watermarked image, $\mathbfit{G}_\text{o}$ is divided into several patches and compared with their corresponding patches from the ground truth image, $\mathbfit{C}$ as shown in Fig. \ref{main_proposed_flowchart}. The discriminator model predicts a score for each of the patch pairs. If the generator is able to perfectly reconstruct the generated watermarked image $\mathbfit{G}_\text{o}$,  keeping it similar to the ground truth image, $\mathbfit{C}$ a higher number of patches predicts probabilities close to 1 (output of well-reconstructed or almost real images is depicted with $\checkmark$). On the other hand, if the generator reconstructs the image poorly, the prediction of the discriminator is close to 0 (output of poor reconstruction or easily detected as generated patches are depicted with  $\times$).

\begin{figure}[h]
    \centering
    \includegraphics[width=1\linewidth]{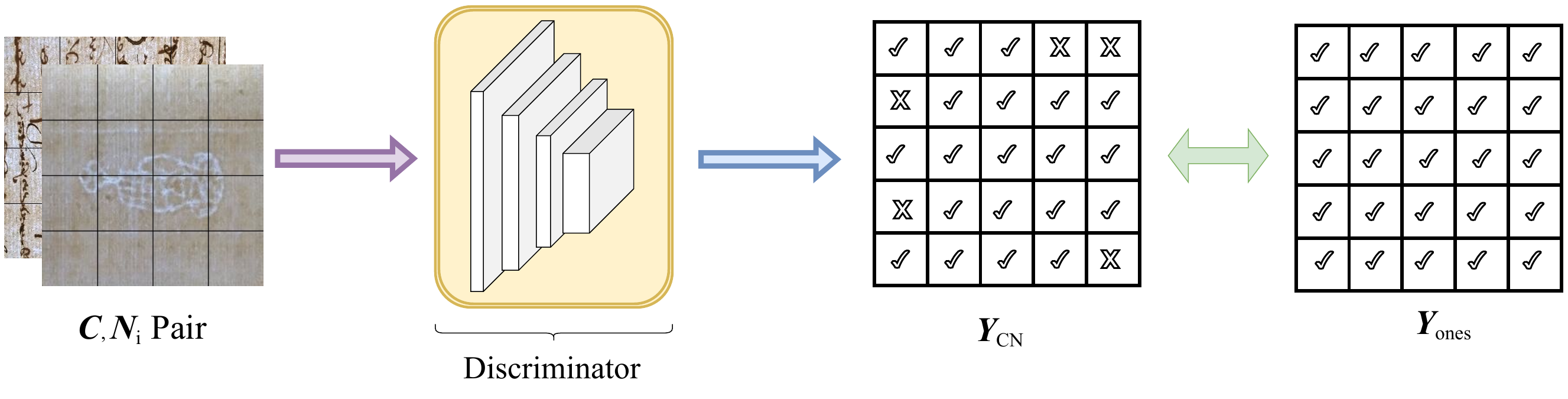}
    \caption{The discriminator of the proposed U-net-based GAN used for extracting clean watermarked image}
    \label{discriminator}
\end{figure}

In order to train the discriminator, firstly score for patches from the pair of noisy watermarked images, $\mathbfit{N}_\text{i}$, and ground truth image, $\mathbfit{C}$ is calculated. Then prediction of patches from another pair containing the noisy watermarked image, $\mathbfit{N}_\text{i}$, and the generated watermarked image, $\mathbfit{G}_\text{0}$ is obtained as shown in Fig. \ref{main_proposed_flowchart} and Fig. \ref{discriminator}.
\begin{algorithm}
\caption{Training of Npix2Cpix GAN}
\label{alg:refinement_keypts}
    \begin{algorithmic}
    \\
        \State $\mathbfit{N}_\text{i} \gets  \text{Noisy\ Watermarked\ Image}$
        \State $\mathbfit{C} \gets \text{Ground\ Truth\ Image\ (clean\ watermark)}$
        \State $\mathbfit{G}_\text{o} \gets \text{Generated\ Watermarked\
         Image}$
     \\
    \end{algorithmic}
    \textbf{for} the number of training iterations \textbf{do}
     \begin{algorithmic}[H]
     \State a. Train Discriminator:    
    \State \quad \quad $\bullet \ \mathbfit{G}_\text{o} = \text{Generator}(\mathbfit{N}_\text{i})$
    \State\quad\quad $\bullet$ $\mathbfit{Y_\text{{CN}}}$ = Discriminator($\mathbfit{C}$, $\mathbfit{N}_\text{i}$)
    \State\quad\quad $\bullet$ $\text{Loss}_{\text{D}_\text{{real}}}$ = MSE($\mathbfit{Y_\text{{CN}}}$, $\mathbfit{Y_\text{{ones}}}$)
    \State\quad\quad $\bullet$ $\mathbfit{Y_\text{{GN}}}$ = Discriminator($\mathbfit{G}_\text{o}$, $\mathbfit{N}_\text{i}$)
     \State\quad\quad $\bullet$ $\text{Loss}_{\text{D}_\text{gen}}$ = MSE($\mathbfit{Y}_\text{{GN}}$, $\mathbfit{Y}_\text{{zeros}}$)
    \State\quad\quad $\bullet$ $\mathrm{\text{Loss}_\text{D}}$ = ($\text{Loss}_{\text{D}_\text{real}} + \text{Loss}_{\text{D}_\text{gen}}$)/ $2$
     \State Backpropagate and update Discriminator parameters
    \State b. Train Generator:
    \State\quad\quad $\bullet$ $\text{Loss}_{\text{G}_\text{real}}$ = MSE($\mathbfit{Y}_\text{{CN}}$, $\mathbfit{Y}_\text{{zeros}}$)
    \State\quad\quad $\bullet$ $\text{Loss}_{\text{G}_\text{gen}}$ = L1($\mathbfit{G}_\text{o}$, $\mathbfit{C}$)
    \State\quad\quad $\bullet$ $\text{Loss}_\text{G}$ = ($\text{Loss}_{\text{G}_\text{real}}  + \text{Loss}_{\text{G}_\text{gen}}$)/ $2$
    \State Backpropagate and update discriminator parameters
  \end{algorithmic}
\textbf{end}
\end{algorithm}

As the main intention behind the discriminator is to differentiate between the ground truth image, $\mathbfit{C}$ from the dataset and the generated watermarked image, $\mathbfit{G}_\text{o}$ from the generator, the discriminator has to emphasize how dissimilar $\mathbfit{C}$ is from the noisy watermarked image, $\mathbfit{N}_\text{i}$ as well as how similar the generated watermarked image, $\mathbfit{G}_\text{o}$ is to noisy watermarked image, $\mathbfit{N}_\text{i}$. As mentioned in Algorithm \ref{alg:refinement_keypts}, a ground truth label of all ones, $\mathbfit{Y}_\text{ones}$ is provided for the discriminator output, $\mathbfit{Y}_\text{CN}$ from the pair of noisy watermarked image, $\mathbfit{N}_\text{i}$ with ground truth, $\mathbfit{C}$. For the other part of discriminator training,  a ground truth of all zeros, $\mathbfit{Y}_\text{zeros}$ is used to calculate loss against discriminator output, $\mathbfit{Y}_\text{GN}$ from a pair containing generated watermarked image, $\mathbfit{G}_\text{o}$ and noisy watermarked image, $\mathbfit{N}_\text{i}$. Thus, the discriminator learns to predict which images are original and which are generated. 

The discriminator loss, $\text{Loss}_\text{{D}}$ is the combination of two different losses, mean square error (MSE) \cite{pishro2016introduction} calculated from, $\mathbfit{Y}_\text{CN}$ and $\mathbfit{Y}_\text{GN}$ calculated from their corresponding ground truths. After obtaining patch-wise predictions from the discriminator model, discriminator generated loss, $\text{Loss}_{\text{D}_\text{gen}}$ and discriminator real loss, $\text{Loss}_{\text{D}_\text{real}}$ calculated for both pairs. As mentioned before, the goal of the discriminator is to discriminate between real and generated images and predict correctly. The intuition behind these losses is that optimizing $\text{Loss}_{\text{D}_\text{real}}$  will teach the discriminator to recognize real images and predict 1 for them, whereas optimizing $\text{Loss}_{\text{D}_\text{gen}}$ will teach the discriminator to recognize generated images and predict 0 for them. The total discriminator loss, $\text{Loss}_{\text{D}}$ is the weighted mean of $\text{Loss}_{\text{D}_\text{{real}}}$ and $\text{Loss}_{\text{D}_\text{gen}}$. So, the discriminator loss $\text{Loss}_\text{D}$ can be written as Eq. (\ref{eqn: dis_loss}).

\begin{equation}
\label{eqn: dis_loss}
  \text{Loss}_\text{D} = \frac{w_{\text{D}_\text{real}}\cdot\text{Loss}_{\text{D}_\text{real}} + w_{\text{D}_\text{gen}}\cdot\text{Loss}_{\text{D}_\text{gen}}}{2}  
\end{equation}
Here, $w_{\text{D}_\text{real}}$ is the discriminator generated loss weight and $w_{\text{D}_\text{gen}}$ is the discriminator real loss weight. 

The main target of the generator is to reconstruct the noisy watermarked image, $\mathbfit{N}_\text{i}$  in such a way that the generated watermarked image, $\mathbfit{G}_\text{o}$ is close enough to the ground truth image, $\mathbfit{C}$, as well as confuse the discriminator. That is why, here the generator loss,  $\text{Loss}_\text{G}$  is calculated from MSE of discriminator's output from the pair of $\mathbfit{C}$ and $\mathbfit{N}_\text{i}$ against $\mathbfit{Y}_\text{zeros}$ and L1 loss \cite{willmott2005advantages} calculated between $\mathbfit{G}_\textit{o}$ and $\mathbfit{C}$. These losses, denoted by $\text{Loss}_{\text{G}_\text{real}}$ and $\text{Loss}_{\text{G}_\text{gen}}$, respectively, are combined to obtain  $\text{Loss}_{\text{G}}$. Here, $\text{Loss}_{\text{G}_\text{gen}}$ fulfills the goal of the generator by directly comparing the generated watermarked image, $\mathbfit{G}_\textit{o}$ with ground truth image,  $\mathbfit{C}$ and $\text{Loss}_{\text{G}_{\text{real}}}$ does it by confusing the discriminator through optimizing in the opposite direction of the gradient updated at discriminator training. Here, gradients are updated in the opposite direction of $\text{Loss}_{\text{D}_\text{real}}$. So, the generator loss, $\text{Loss}_\text{G}$ can be described as Eq. (\ref{eqn: gen_loss})

\begin{equation}
  \label{eqn: gen_loss}
  \text{Loss}_\text{G} = \frac{w_{\text{G}_\text{real}}\cdot\text{Loss}_{\text{G}_\text{real}} + w_{\text{G}_\text{gen}}\cdot\text{Loss}_{\text{G}_\text{gen}}}{2}  
\end{equation}
Here, $w_{\text{G}_\text{real}}$ is the generator's generated loss weight, and $w_{\text{G}_\text{gen}}$ is the generator's real loss weight. 

In our approach to denoising noisy historical documents using the proposed Npix2Cpix GAN, we have chosen to keep the weights for both real and generated losses equal. This decision is based on the need to maintain balance in GAN training. By keeping the weights equal, we ensure that both real and generated losses contribute equally, preventing either the generator or the discriminator from becoming too dominant. This balance is crucial for achieving stable and effective training dynamics.

In order to improve the denoising performance of the proposed Npix2Cpix GAN, it is important to concurrently minimize both $\text{Loss}_\text{{G}}$ and $\text{Loss}_\text{D}$ by optimizing the network parameters. Any imbalance in training the generator and discriminator sometimes leads the training process to diverge or collapse. Upon successful training, the generator of the trained GAN is expected to produce a denoised watermarked image from a degraded historical watermarked image. 
This denoised watermarked image is then utilized in Algorithm \ref{alg: Siamese} for classifying the exact class of watermarks which is explained elaborately in section \ref{ssec:c}.
\subsection{Metric-Based Classifier: Siamese Network}
\label{ssec:c} 
In the subsequent phase of our proposed work, a Siamese-based one-shot classification network is introduced for categorizing denoised historical watermarked documents. Siamese networks are, as the name suggests, a two-layered architecture. Instead of employing classification loss functions to train the model to categorize its input data, this network trains the model to differentiate between two sets of data. It analyzes two sets of data using a similarity (or distance) metric in order to identify whether they belong to the same class. This model comprises two identical neural networks, also known as feature extractors, with a single input data point and shared parameters as depicted in Fig. \ref{fig:Siamese}. Since the weights and biases are the same throughout, identical types of features are retrieved in the middle.

In this paper, the challenges associated with classifying noisy watermarks are clearly outlined in subsection \ref{ssec:ps}, which are comprehensively addressed through the proposed conditional U-Net-based GAN. However, there may still arise some challenging scenarios where documents remain significantly degraded and may not be fully denoised. As a result, efforts are focused on improving the feature extraction functionality of the Siamese network so that important features can be extracted from marginally denoised watermarked images. In order to accomplish this, adjustments are implemented to the conventional Siamese network, enabling the smooth incorporation of diverse pretrained networks such as ResNet, MobileNet, and EfficientNet for the purpose of extracting meaningful features. In addition, to enhance the discriminatory capabilities of the network, the cosine distance is utilized to compute the similarity between image pairs, instead of the conventional Euclidean distance.

\begin{algorithm}
\caption{Training of Siamese Network}
\label{alg: Siamese}
    \begin{algorithmic}
    \\
        \State $\mathbfit{X}_1 \gets  \text{Image 1}$
        \State $\mathbfit{X}_2 \gets \text{Image 2}$
        \State $y \gets \text{Similar\ or\ Dissimilar\ label}$
        \\
        \State $m \gets \text{Expected\ Margin}$
        \State $\text{Net} \gets \text{Siamese\ Backbone\ Feature\ Extractor}$
        
        \State $\text{Dist} \gets \text{Distance\ Metric}$
     \\
    \end{algorithmic}
    
    \textbf{for} the number of training iterations \textbf{do}
     \begin{algorithmic}[H]
    \State $\bullet$ $\mathbfit{Y}_1$ = Net($\mathbfit{X}_1$)
    \State $\bullet$ $\mathbfit{Y}_2$ = Net($\mathbfit{X}_2$)
    \\
    \State $\bullet$ $D_\text{c}$ = Dist($\mathbfit{Y}_1$, $\mathbfit{Y}_2$)
    \State $\bullet$ $\text{Loss}_\text{S}$ = $(1-y)D_\text{c}^2 + y\{\text{max}(0,m-D_\text{c})\}^2$ 
    \State $\bullet$ Backpropagate and update parameters to reduce loss
      
  \end{algorithmic}
\textbf{end}
\end{algorithm}

The architecture of our implementation for Siamese-based one-shot classification consists of three components: a feature extractor, a feature comparison metric, and a suitable loss function. At first, image features $\mathbfit{Y}_\text{1}$ and $\mathbfit{Y}_\text{2}$ are extracted from input images $\mathbfit{X}_\text{1}$ and $\mathbfit{X}_\text{2}$ respectively after the images are cleaned with the proposed Npix2Cpix GAN. The methodology for the proposed conditional GAN-based denoising is covered in detail in subsection \ref{npix2cpix}. The distance, $D_\text{c}$ between these two features, is calculated using the distance metric as shown in Fig \ref{fig:Siamese}. The employed unit of measurement is the cosine similarity \cite{lahitani2016cosine}. The ground truth label 0 is used if the image pair is taken from the same class and a label of 1 is given if images from the pair are from different classes, which is denoted by $y$. When trained with these labels, the model gains the ability to distinguish between features belonging to different classes and to concentrate on watermark similarities.

To train the model for detecting the similarity, we need a loss function that can work with the distance, $D_\text{c}$ calculated from extracted features. The loss function has to minimize $D_\text{c}$ of the input pair from the same class and maximize $D_\text{c}$ up to a margin, $m$ if the input images are from a different class, as depicted in Algorithm \ref{alg: Siamese}. The Contrastive loss, a differentiable loss function, is used as Siamese loss, as it can fulfill this requirement. Our implementation of the loss function, $\text{Loss}_\text{S}$ is denoted by Eq. (\ref{eqn:loss_sia})
\begin{equation}
  \label{eqn:loss_sia}
  \mathrm{\text{Siamese loss},  \text{Loss}_\text{S}} = (1-y)D_\text{c}^2 + y\{\text{max}(0,m-D_\text{c})\}^2   
\end{equation}

The final prediction of the model is represented here by $D_\text{c}$, the cosine distance between two extracted features. The dataset label is denoted by the value $y$, and the margin value, $m$ represents the expected variation in pair distance scores between similar and dissimilar pairs.

Two components compose the loss function. While the label value for dissimilar images is set to 1, it is set to 0 for similar images. For similar image combinations, $y$ is 0, giving the second half of the loss equation zero. The first part will thus be prioritized in the loss minimization, and the network will learn to predict similar features for similar class images. On the other hand, in the case of dissimilar images with $y$ close to 1, the second part of the loss function $\text{Loss}_\text{S}$ will get priority, and will be minimized while the first half vanishes. In order to minimize this part, it is intended that 0 is always chosen from the max function. For this to happen, the other argument $ (m - D_\text{c}) $ has to become negative, which means the model has to predict a value of $D_\text{c}$ that is larger than the margin. This is how the model is trained to recognize similar and dissimilar pairs.

\subsubsection{Training Procedure of Siamese Network}
First, our suggested U-net-based conditional GAN is used to denoise all the impaired images from 100 classes (50 images/class). The cleaned images of 100 classes are then split into a 60:40 ratio for the training and inference procedure.  During the training phase, the model is trained using a similarity learning strategy. In order to do this, it is necessary to use image pairings from similar and dissimilar classes. These pairs are produced using the cleaned 60\% (60 classes) data of the 'A Classification'  dataset. A randomly selected class from the train portion data is picked, and two randomly selected images from that class are taken to create similar pairings. Again two classes are arbitrarily selected for the dissimilar pairs, and then one image from each class is selected for the pair. 1000 additional image pairings are used for validation after each 10,000 training pairs for each epoch.

 The feature extractor is utilized to extract each image's significant features. Finally, the distance metric is used to estimate the feature distance score, which is then compared to the dataset label. Over time, the model develops the ability to compare watermarks and distinguish them from background noise or handwriting obstruction.

\subsubsection{Inference Procedure of Siamese Network}
One support set image appears for each class during the one-shot learning inference. Using these, the target set (test) images must be categorized into the classes each support set image stands for.

After the training phase, the model learned to forecast the distance score of features from each pair of images. The proposed approach looks for the image with the lowest distance score since the objective is to find the image that most closely resembles the target set image. To do this, the support set class that closely matches the target image is determined by comparing each target set image to all the support set images. Here, the original clean images of each class are utilized as support set images rather than the generated clean image.

\subsubsection{One-shot classification}
For the purpose of classification, one-shot classification with a Siamese network is used. The difficult issue of historical watermark categorization is addressed using a one-shot learning-based classification technique. The one-shot test shows that the proposed GAN-based technique can denoise the watermarked documents in the majority of situations when genuine clean watermarked images are used as the support set and generated watermarked images are used as the target set for classification. To distinguish watermarks from dissimilar classes in the compressed feature space, the suggested architecture is built on a Siamese network with several backbones (ResNet, MobileNet, and EfficientNet).

\subsection{An Additional Experiment for U-net}
\begin{figure}[h!]
    \centering
    \includegraphics[width=0.7\linewidth]{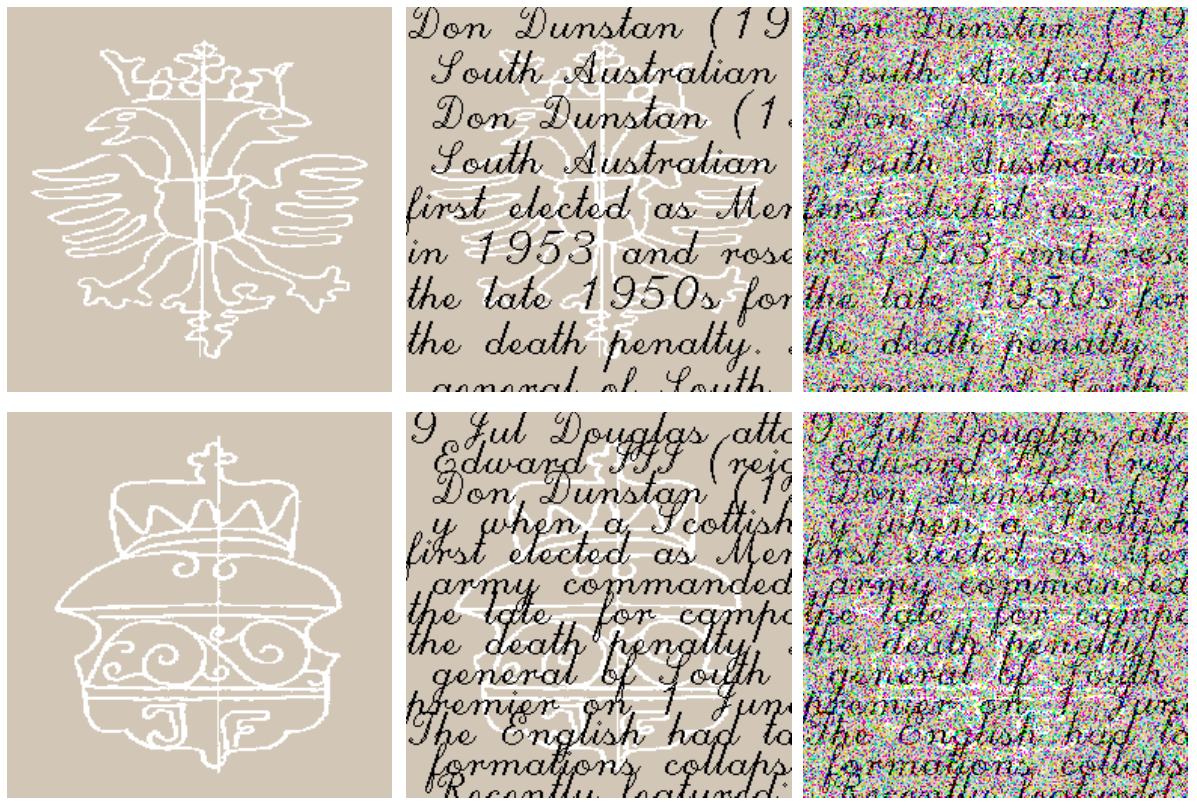}
    \caption{These are a few examples from the synthetic dataset, where the left one is the ground truth, and the middle one is a degraded image with handwriting. Adding Gaussian noise over the handwriting for complexity indicates the right one}
    \label{triplet_syn}
\end{figure}

\begin{figure}[h]
    \centering
    \includegraphics[width=1\linewidth]{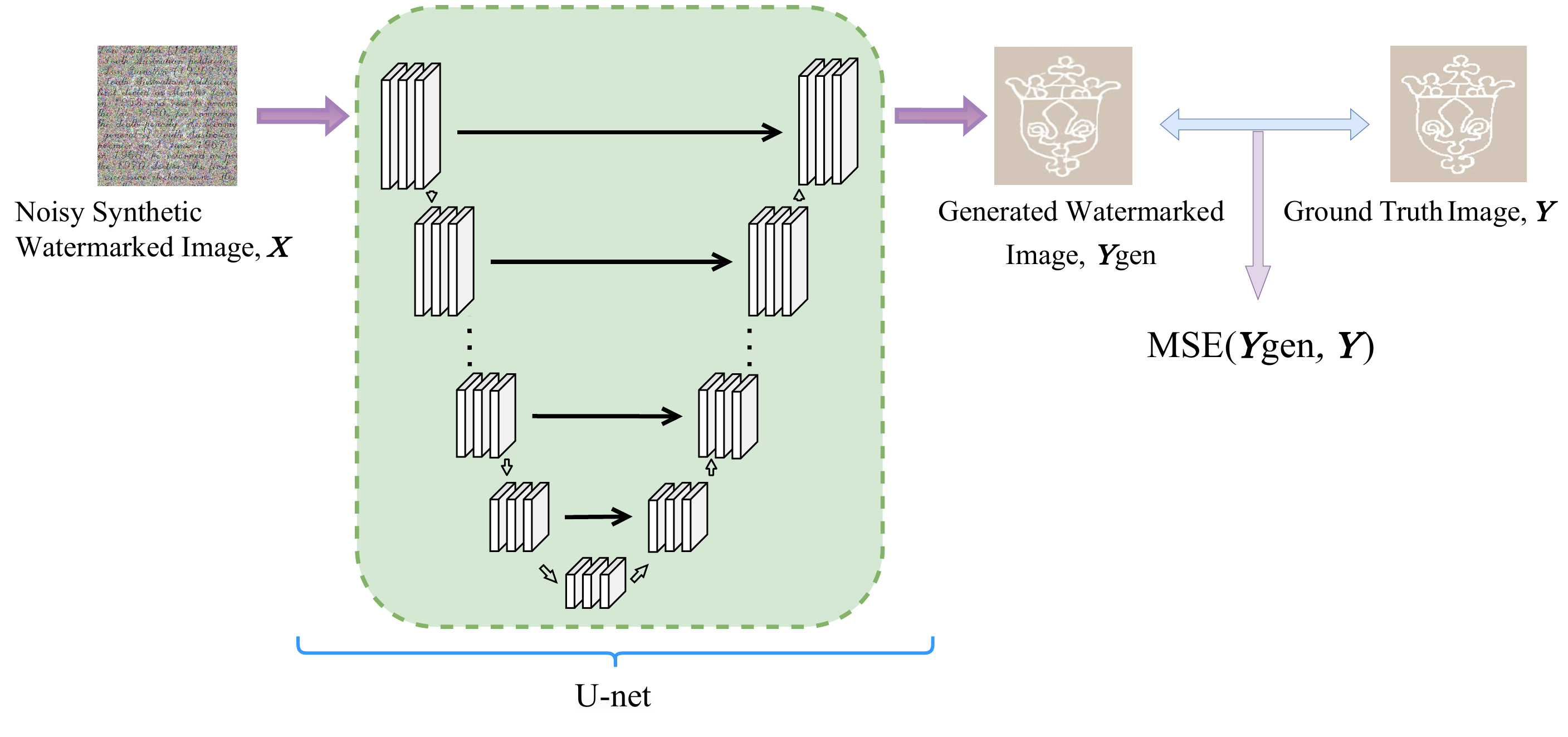}
    \caption{The generator of the proposed U-net-based GAN used for extracting clean watermarks from synthetically generated images}
    \label{only_unet}
\end{figure}

Next, the performance of the proposed U-net-based design for the generator model is investigated. In Fig. \ref{only_unet}, with the help of a simplified block diagram, the task of extracting clean watermarks from the synthetically generated documents is explained using the generator of the proposed U-net-based GAN. Synthetic documents that have been degraded by the addition of random handwriting are used to estimate how well the suggested model performs. Moreover,  random Gaussian noise is also incorporated into the document, as shown in Fig. \ref{triplet_syn}, to make the task more challenging.

Fig. \ref{only_unet} depicts the generated watermarked synthetic image, $\mathbfit{Y}_\text{gen}$, which is quite similar to the ground truth image, $\mathbfit{Y}$, where, $\mathbfit{X}$ is the synthetic watermarked image with handwriting and noise. This experiment is carried out to test whether the proposed U-net-based generator is robust enough to denoise the document which belongs to a different domain from the noise-corrupted synthetic document and retrieve the watermark.  

\section{Result and Analysis}
In this section, experimental details of the proposed work and comparative performance analysis with existing methods are presented. The training of the proposed GAN-based model is conducted on a Tesla P100 GPU, requiring approximately 10 hours to complete. Subsequently, for the Siamese-based one-shot classification task, the same Tesla P100 GPU is employed, with a training duration of 4.5 hours. The experimental setup utilized a single Tesla P100 GPU with 3584 CUDA cores and 16GB GDDR6 VRAM, a single-core hyper-threaded Xeon processor at 2.2GHz, 15.26GB of RAM, and 155GB of available disk space. It is important to note that the time required for specific tasks may vary depending on the number of classes involved. In the context of this study, the reported time duration pertains to experiments conducted with the dataset comprising 100 classes. Additionally, the training configurations and hyperparameters used for training the proposed U-net-based GAN and Siamese network are described in Table \ref{tab:hyper}.

\begin{table}[t]
\centering
\small
\caption{Specific configurations and hyperparameters used for training}
\label{tab:hyper} 
\begin{tabular}{l c c} 
\hline
Hyperparameters & Npix2Cpix & Siamese network\\ [0.5ex] 
\hline
Batch size & 16 & 8 \\
Learning rate & 0.0002 & 0.001 \\
Betas & [0.5, 0.999] & [0.9, 0.999] \\ 
Optimizer & Adam & Adam \\
Epochs & 200 & 35 \\
Image size & $256\times256$ & $256\times256$ \\
\hline
\end{tabular}
\end{table}

A few metrics \cite{sara2019image}, that are used to evaluate how precisely the watermark is extracted, are demonstrated to prove the viability of the suggested technique. In Fig. \ref{qualitative_real}, a generated image from the noisy sample, which is quite similar to the ground truth, is shown. Here, the extracted watermark is marked by a green box for better visualization. The table contains a few parametric metrics such as MSE, PSNR, SSIM, and BRISQUE.

\begin{description}
    \item[MSE:] Mean squared error (MSE) is the most widely used estimator of an image quality metric. Lower values of MSE are better. MSE is described by Eq. (\ref{eqn:mse}).
    \begin{equation}
    \label{eqn:mse}
      \text{MSE} = \frac{1}{n}\sum_{i=1}^{n}\left( \mathbfit{Y}_\textit{i} - \hat{\mathbfit{Y}_\textit{i}} \right)  
    \end{equation}
    where $\mathbfit{Y}_\textit{i}$ is the ground truth image and $\hat{\mathbfit{Y}_\textit{i}}$ is the generated image.
    \item[PSNR:] The peak signal-to-noise ratio (PSNR) between two images is used to compare the original picture quality to the transformed image quality. The PSNR increases with the quality of the compressed or reconstructed image. PSNR and MSE are used to measure the quality of image compression. The mathematical expression of PSNR is shown in Eq. (\ref{eqn:psnr}). 
    \begin{equation}
    \label{eqn:psnr}
       \text{PSNR} = 10\ \text{log}_{10} \left( \frac{\mathrm{{\text{MAX}_I}^2}}{\text{MSE}} \right) 
    \end{equation}
     
    where $\mathrm{\text{MAX}_I}$ is the maximum possible pixel value of the image.
    
    \item[SSIM:] In order to measure the similarity between two digital images or videos, the structural similarity index metric (SSIM) is used \cite{wang2004image}. An original, uncompressed, distortion-free reference image is used to evaluate the quality of the test image. The SSIM index can provide precise information on how closely the generated watermark is to its ground truth.  Between two images $x$ and $y$, the SSIM can be expressed as Eq. (\ref{eqn:ssim}).
    \begin{equation}
    \label{eqn:ssim}
    \text{SSIM} = \frac{\left( 2\mu_x \mu_y + c_1 \right) \left( 2\sigma_{\text{xy}} + c_2 \right)}{\left( \mu_x^2 + \mu_y^2  + c_1\right) \left( \sigma_x^2 + \sigma_y^2  + c_2 \right)}   
    \end{equation}

    where $\mu_x$ and $\mu_y$ are pixel mean values of images $x$ and $y$, $\sigma_x^2$ and $\sigma_y^2$ are their variance respectively, $\sigma_{\text{xy}}$ is covariance of images $x$ and $y$. In order to stabilize the divisions with weak denominators, two variables $c_1$ and $c_2$ are used.

      \item[BRISQUE:] \> \> \> It is a spatially-based distortion-generic no-reference (NR) image quality assessment (IQA) model that utilizes natural scene statistics, instead of distortion-specific features like ringing, blur, or blocking. Blind Image Spatial Quality Evaluator (BRISQUE) \cite{7051498} uses locally normalized luminance coefficients of the given image to estimate potential losses of "naturalness" in the image due to the presence of distortions, providing a comprehensive evaluation of quality.
\end{description}

\subsection{Qualitative evaluation}
\begin{figure}[ht]
    \centering
    \includegraphics[width=1\linewidth]{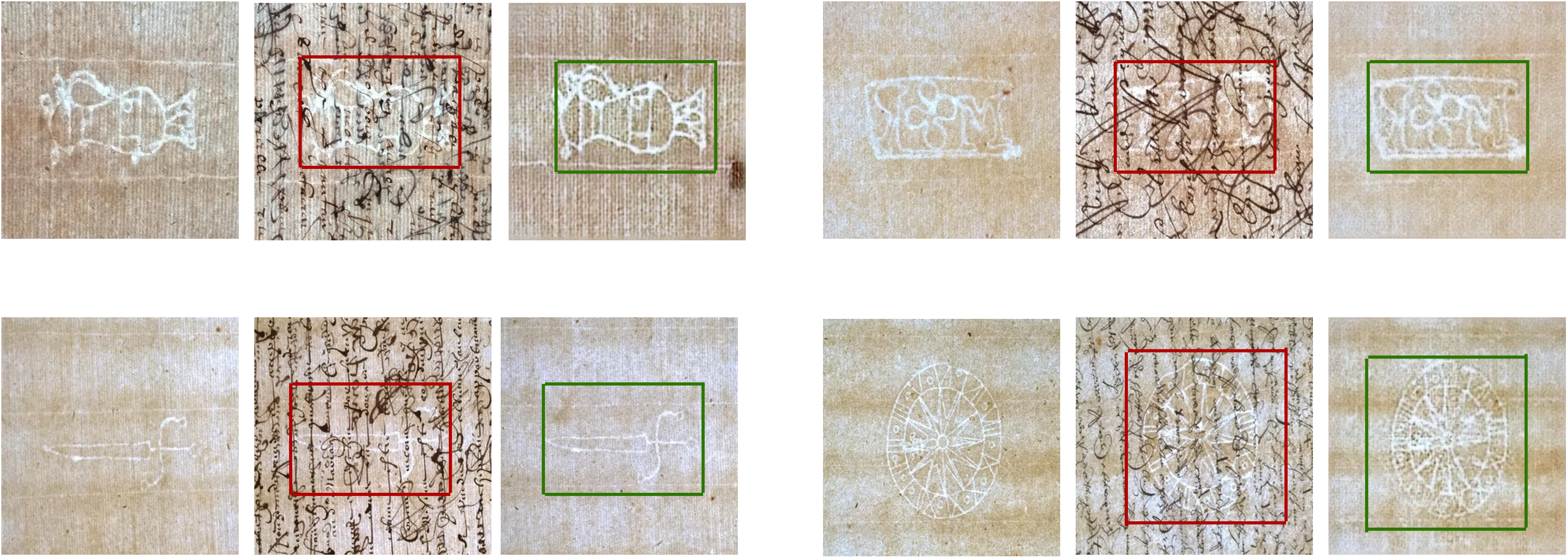}
    \caption{The triplet of four classes output is shown here. The left image indicates the ground truth, where the middle is the noisy image and extracted clean watermarks are on the right side of each triplet}
    \label{qualitative_real}
\end{figure}

Fig. \ref{qualitative_real} presents a qualitative review of our findings. Here, extracted watermarks from four separate classes are displayed. The actual degraded images in the middle have been pre-processed using our proposed U-net-based conditional GAN.
The generated images are very similar to the ground truth images, as shown in Fig. \ref{qualitative_real}. Green boxes indicate the retrieved watermarks, while red boxes indicate the presence of a watermark in degraded historical documents.

\begin{figure}[h]
    \centering
    \includegraphics[width=1\linewidth]{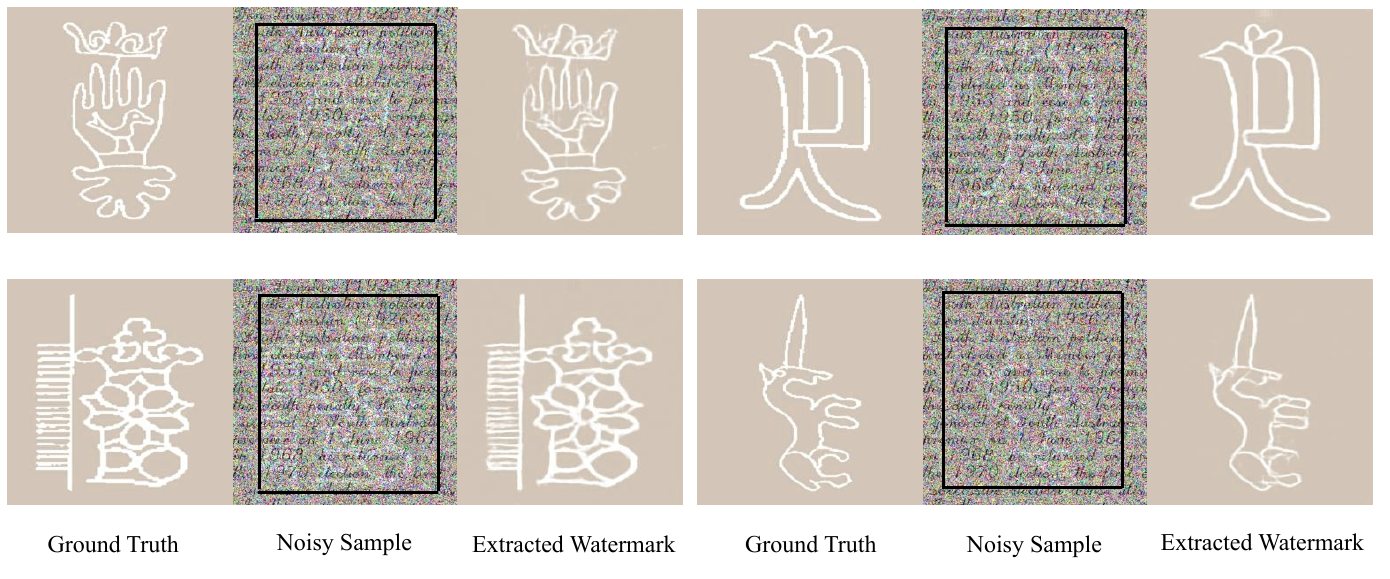}
    \caption{The triplet of four classes output is shown here for the synthetic dataset. The left image indicates the ground truth, where the middle is the artificially degraded image and extracted clean synthetic watermarks are on the right side of each triplet}
    \label{qualitative_syn}
\end{figure}

Similar results can be shown in Fig. \ref{qualitative_syn}, where synthetic images, of a different domain, are artificially deteriorated using handwriting and random Gaussian noise. As the background pixels in these synthetic images are evenly dispersed, in contrast to how they are spread in ancient watermarked images, the denoised watermarks are more obvious in Fig. \ref{qualitative_syn}. 
These findings show that the proposed network, or the generator, is capable of producing statistically rich images from actual and synthetic noisy images presented in a different domain.

\subsection{Quantitative Evaluation}
For the quantitative assessment of the performance of the proposed watermarked image denoising method, initially, a few image quality metrics are computed for both actual noisy images as well as generated images, such as MSE, RMSE, SSIM, and BRISQUE. Results are compared in Table \ref{Quantitative evaluation of generated clean images}. From Table \ref{Quantitative evaluation of generated clean images}, it is quite evident that all quantitative measures are improved when generated images are compared to the actual noisy images.

\begin{table}[h]
\caption{Quantitative evaluation of generated clean images}
\label{Quantitative evaluation of generated clean images}
\centering
\def\arraystretch{1.25}
\resizebox{0.8\textwidth}{!}{
    \begin{tabular}{lcccc}
    \hline
    Split & MSE & RMSE  & PSNR (dB) & SSIM  \\ 
    \hline
    Noisy & 104.567 & 10.225 & $10.225$ & 0.117  \\ 
    Generated Clean & 71.015 & 8.429 & $29.615$ & 0.555  \\ 
    \hline
    \end{tabular}
} 
\end{table}

The results on the selected metrics, in the case of synthetic data, are shown in Table \ref{Quantitative evaluation of generated clean images (synthetic domain)}. Similar performance is also observed here, which provides additional support for the dependability of the proposed architecture in a different domain.

\begin{table}[h]
\caption{Quantitative evaluation of generated clean images (synthetic domain)}
\label{Quantitative evaluation of generated clean images (synthetic domain)}
\centering
\def\arraystretch{1.25}
\resizebox{0.8\textwidth}{!}{
    \begin{tabular}{lcccc}
    \hline
    Split & MSE & RMSE  & PSNR (dB) & SSIM  \\ 
    \hline
    Noisy & 106.709 & 10.33 & $27.849$ & 0.1067  \\
    Generated Clean & 14.971 & 3.869 & $36.378$ & 0.941 \\
    \hline
    \end{tabular}
}
\vspace{8pt}
\end{table}

The BRISQUE score comparison is shown with several images from a definite class in Table \ref{Quantitative evaluation of generated clean images (Brisque Score)}. Typically, the BRISQUE score spans between 0 and 100. The improved perceptual quality of the image is reflected in lower score values.
Table \ref{Quantitative evaluation of generated clean images (Brisque Score)} shows that the generated images using the proposed technique have a score that is rather near to the ground truth images in both domains.

\begin{table}[h]
\caption{Quantitative evaluation of generated clean images (BRISQUE Score)}
\label{Quantitative evaluation of generated clean images (Brisque Score)}

\centering
\def\arraystretch{1.25}

\begin{tabular}{lc}
\hline
Images & BRISQUE  \\
\hline

Ground Truth Image & 23.324 \\
Noisy Watermarked Image & 30.518  \\
Generated Watermarked Image & 24.139  \\

Ground Truth Image (Syn.) & 56.271   \\
Noisy Watermarked  Image (Syn.) & 88.663  \\
Generated Watermarked Image (Syn.) & 54.033  \\
\hline

\end{tabular}
\vspace{8pt}

\end{table}

The watermark classification accuracy achieved through our proposed method is compared with some recent methods dealing with the same problem and results are shown in Table \ref{gan_withoutgan}.

\begin{table}[h]
\caption{Performance comparison with different methods}
\label{gan_withoutgan}
\centering
\def\arraystretch{1.25}

\resizebox{0.8\textwidth}{!}{
\begin{tabular}{lc}
\hline
Method & Accuracy  \\
\hline

Local Spatially-Aware Approach \cite{shen2021large}, \cite{shen2022spatially} & 81.2\% \\
One-shot + Thresholding-based Enhancement \cite{mypaper} & 80\% \\
Proposed & 95\%  \\

\hline

\end{tabular}
}
\vspace{8pt}

\end{table}

It is observed from Table \ref{gan_withoutgan} that, the proposed method achieves 95\% classification accuracy, which is much higher than the accuracies obtained by some existing works such as the local spatially-aware approach and Siamese network with thresholding-based enhancement. So, it can be concluded that the proposed one-shot classification with a GAN-based pre-processing is more efficient than the existing state-of-the-art methods.

\subsection{Ablation Experiment}
\subsubsection{Effect of Npix2Cpix GAN}
To show the effectiveness of the proposed GAN-based denoising, we conducted experiments to compare the performance of the one-shot approach with and without GAN-based denoising prior to the classification stage. The results of these experiments are detailed and compared in Table \ref{abl:gan-siam}.
\begin{table}[h]
\caption{Classification Performance comparison with/without Npix2Cpix}
\label{abl:gan-siam}
\centering
\def\arraystretch{1.25}

\resizebox{0.75\textwidth}{!}{
\begin{tabular}{lc}
\hline
Method & Accuracy  \\
\hline
Siamese Network & 76.2\% \\
Proposed Npix2Cpix + Siamese Network & 95\%  \\

\hline

\end{tabular}
}
\vspace{8pt}

\end{table}
Table \ref{abl:gan-siam} clearly shows that the Siamese network with the proposed U-Net-based conditional GAN performs better. In particular, performance is noticeably degraded when the proposed GAN-based denoising method has not been utilized. This is because the feature extractor of the Siamese network has to work with historical watermarked images that are severely distorted for classification. However, using the proposed method to remove noise from the historical watermarked document leads to a major boost in classification accuracy, since the feature extractor now has to handle a noise-free watermarked image.
\subsubsection{Effect of different feature extractor of Siamese network}
For the purpose of classifying the denoised watermarked image, a Siamese-based one-shot method is demonstrated in this work. As previously discussed, the feature extractor is constructed in such a manner that the backbone network can be simply altered for the one-shot method experiment. Results using several backbones are shown in Table \ref{0ne-shot}, which shows the classification accuracy in the one-shot approach. The findings for the feature extractors such as Resnet18 \cite{he2016deep}, Efficientnetb0 
\cite{tan2019efficientnet}, and MobilenetV3 (small) \cite{howard2019searching} are presented in Table \ref{0ne-shot}, with Resnet18 architecture providing the best one-shot test accuracy (95\%). Due to the extensive use of parameters in this design, Resnet18 is employed here, which outperforms the other architectures. Efficientnet is employed because of its ability to optimize both parameters and performance. Finally, one-shot performance with fewer parameters is checked using the MobilenetV3 (small) results. 

\begin{table}[h]
\caption{One-shot classification results for different backbones}
\label{0ne-shot}

\centering
\def\arraystretch{1.25}

\resizebox{0.9\textwidth}{!}{
\begin{tabular}{lcc}
\hline
Backbone & Numbers of Parameters & One-Shot Best Accuracy   \\
\hline

Resnet18 & 11.2M & 95\%  \\

Efficientnetb0 & 4.08M & 86.3\%   \\

MobilenetV3 (small) & 0.62M & 85\%  \\

\hline

\end{tabular}
}
\vspace{8pt}

\end{table}


\begin{table}[!h]
\caption{Performance comparison in different scales (\% accuracies)}
\label{tab:scale}

\centering
\def\arraystretch{1}

\resizebox{0.9\textwidth}{!}{
\begin{tabular}{lcccccc}
\hline
Architecture & Scale-1 & Scale-2  & Scale-3 & Scale-4 & Scale-5 \\
\hline

MobilenetV3 (small) & 70\% & 85\% & 78.7\% & 77.5\% & 81.2\% \\

Efficientnetb0 & 68.8\% & 75\% & 78.7\% & 86.3\% & 83.8\%  \\

Resnet18 & 91.2\% & 87.5\% & 95\% & 91.2\% & 86.3\%\\
\hline

\end{tabular}
}
\vspace{8pt}

\end{table}

\subsubsection{Effect of image shape variations}
For experimentation with the proposed one-shot classification method images are resized to 5 scales. In Table \ref{tab:scale}, results of classification accuracies obtained for several scales are presented. Scale-\textit{i} corresponds to the specific size $192\times192$, $224\times224$, $256\times256$, $352\times352$, $400\times400$, respectively for \textit{i} = 1,...,5.
In all the scales, the Resnet18 backbone provides consistently better classification accuracy and exhibits relatively less variation with respect to scales.  

\section{Limitations and Future Work}
This paper presents a modified U-net-based conditional generative adversarial image-to-image translation network for watermark retrieval from historical documents and a Siamese-based one-shot approach for classifying the retrieved watermarks. The proposed method is validated using a large-scale dataset containing ancient watermarked images. While the method shows promise for denoising noisy watermarked documents, there are a few limitations. Historical documents often exhibit varying degrees of noise, including severe degradation and complex clutter, which may impact the model's performance. In extreme cases, the model may struggle to effectively denoise images, potentially leading to incomplete restoration of watermarks. Additionally, the diversity and complexity of watermarks pose a challenge to the model's scalability. Future research should focus on developing more robust denoising techniques incorporating additional pre-processing steps to handle variations in noise levels. Expanding the training dataset to include more diverse watermark samples from different domains and employing advanced feature extraction methods could further enhance the model's effectiveness.
\section{Conclusion}
This research demonstrates how, despite the fact that watermarks are represented in a variety of ways, deep learning-based algorithms are highly useful for identifying and extracting watermarks from historical document images. This work proposes a U-net-based conditional GAN that can clean noisy historical watermarked documents and retrieve the specific watermarked image. The main difference between the proposed studies and existing ones is the utilization of a generative adversarial network (GAN)-based image-to-image translation method for watermark extraction, followed by a Siamese-based one-shot classification approach, which overcomes issues associated with traditional denoising and classification methods while addressing the scarcity of data. The proposed network is further tested using images from a different domain, and in this instance, rich-quality watermarks are also retrieved. Several metrics and one-shot classification accuracy are used to evaluate the quality of denoised watermarked images. Extensive experiments show that the proposed technique outperforms some recent methods when it comes to classifying the denoised watermarked image utilizing a one-shot approach.

 \bibliographystyle{elsarticle-num} 
 \bibliography{cas-refs}





\end{document}